\documentclass[review]{elsarticle}

\usepackage{lineno,hyperref}
\modulolinenumbers[5]

\journal{Neural Networks}

\usepackage[utf8]{inputenc} 
\usepackage[T1]{fontenc}    
\usepackage{hyperref}       
\usepackage{url}            
\usepackage{booktabs}       
\usepackage{amsfonts}       
\usepackage{nicefrac}       
\usepackage{microtype}      
\usepackage{xcolor}         
\usepackage{graphicx}
\usepackage{xcolor}

\begin{document}
\begin{frontmatter}

\title{All by Myself: Learning Individualized Competitive Behaviour with a Contrastive Reinforcement Learning optimization}

\author{Pablo Barros\\Alessandra Sciutti}
\address{ E-mail: pablo.alvesdebarros@iit.it,
alessandra.sciutti@iit.it\\ CONTACT Unit, Italian Institute of Technology, Italy}

\begin{abstract}
\textcolor{black}{
 In a competitive game scenario, a set of agents have to learn decisions that maximize their goals and minimize their adversaries' goals at the same time. Besides dealing with the increased dynamics of the scenarios due to the opponents' actions, they usually have to understand how to overcome the opponent's strategies. Most of the common solutions, usually based on continual learning or centralized multi-agent experiences, however, do not allow the development of personalized strategies to face individual opponents. In this paper, we propose a novel model composed of three neural layers that learn a representation of a competitive game, learn how to map the strategy of specific opponents, and how to disrupt them. The entire model is trained online, using a composed loss based on a contrastive optimization, to learn competitive and multiplayer games. We evaluate our model on a pokemon duel scenario and the four-player competitive Chef's Hat card game. Our experiments demonstrate that our model achieves better performance when playing against offline, online, and competitive-specific models, in particular when playing against the same opponent multiple times. We also present a discussion on the impact of our model, in particular on how well it deals with  on specific strategy learning for each of the two scenarios. }

\end{abstract}
\end{frontmatter}

\section{Introduction}

\textcolor{black}{
Learning to play competitive games is one of the most challenging scenarios for current reinforcement learning research. Besides having to map a set of states and action pairs in order to learn a specific goal to achieve a successful state, in a competitive scenario an agent has also to predict - and in most cases prevent - actions from its adversaries. In these scenarios, the need to learn conceptual properties from the state transitions is extremely important, as the dynamics of the environment are constantly changing. In particular, the understanding of recurrent patterns in personalized strategies would allow an opponent to adapt towards different opponents weather they are similar or not. Also, understanding the impact of the opponent's actions, and deriving strategies to mitigate them, increases the challenges of achieving an optimal learning scheme. }

\textcolor{black}{
Most of the common solutions for learning competitive games are not easily capable of dealing with an important factor of these scenarios: adaptation towards specific opponents. The majority of the proposed models rely on thousands of learning epochs to achieve super-human performance by learning a general strategy that overcome any adversary \cite{lanctot2017unified, silver2018general, torrado2018deep}. We hypothesize that a model that can adapt its game style against individual opponents, identifying recurrent strategies preferentially during while playing the game, would obtain a better performance against a number of opponents. Ideally, such an agent would also not lose its general capabilities of playing the game against different opponents, avoiding typical transfer learning problems such as catastrophic forgetting.}

In this paper, we propose a mechanism to learn how to adapt to specific adversaries while playing competitive games through a contrastive optimization \cite{chen2020simple} that contributes to general strategy learning. Our contrastive learning model learns how to represent patterns on the game strategy of individual opponents, and uses this knowledge to derive \textcolor{black}{ an auxiliary goal based on the prediction of the opponent's strategy that focuses on disrupting the opponent's game-play style}. Our model combines global and opponent-specific local policy networks that use the contrastive-learned representation to mitigate the chances of the opponent of performing a successful move.

We evaluate our model on two conceptually different competitive learning scenarios: First, we want to assess its capability to adapt against a single opponent in a high-dimensional and dynamic space. Thus, we train and evaluate it using the PokEnv Pokemon battle simulator \cite{poke_env_2020_11_14}. In this scenario, an expert opponent has to learn a general game style that encapsulates all the possible combinations of pokemon and moves given by a random battle initialization. Our goal is to assess how our model can counter such an expert opponent.

In our second evaluation scenario, we assess how our model learns to play the four-players Chef's Hat Card game \cite{barros2020learning}. In this game, the players use very specific rules to discard a set of cards at hand using one of the 200 possible actions per turn. In this scenario, an expert opponent is the one that learns a specific strategy that optimizes its chances of winning taking into account the restricted game rules and the actions of three other opponents. We evaluate our proposed model by measuring how it can mitigate the opponents' chances of winning the game. 

In both scenarios, we perform a series of experiments to fine-tune and provide an ablation understanding of the impact of our experimental decisions on our model. Also, we compare our model's performance it with the most successful reinforcement learning algorithms, including Deep Q-Learning (DQL) \cite{mnih2013playing}, Proximal Policy Optimization (PPO) \cite{schulman2017proximal}, Actor-Critic with Experience Replay (ACER) \cite{wang2016sample}, and Generative Adversarial Imitation Learning (GAIL) \cite{ho2016generative}). Each of these algorithms is adapted to be used in offline and online learning strategies including self-play and transfer learning through fine-tuning. We also compare our solution to specific competitive learning models, DRON \cite{he2016opponent} and Bayes-ToMoP \cite{yang2018towards}. 

\textcolor{black}{ Our experiments demonstrate that besides achieving a performance improvement, our adversary-focused contrastive reinforcement learning leverages its own experience when playing with the different opponents and can rapidly adapt against them. We discussed our results by detailing how our learning strategy impacts the modeling of offline and online learning adversaries, and show that our model is able to adapt towards different types of strategies using very few game rounds. We also discuss the capability of our model to deal with catastrophic forgetting, in particular when playing against different adversaries for longer periods. 
Besides this introduction, this paper includes a related work on Section 2, where we discuss the important literature to understand our contribution; our proposed model on Section 3; the entire experimental setup, including the environment settings and all the implementation details of our models, on Section 4; We report all the results on Section 5, and discuss them on section 6; and finally, we conclude the paper on Section 7. }



\section{Related Work}


Most of the current successful solutions for learning competitive games fall into four learning strategy categories:  imitation learning, transfer learning, competitive learning, and continual learning.




\subsection{The Transfer and Imitation Learning Approach.} 
The problem of adaptation towards novel scenarios is somehow leveraged by the recent 
transfer learning research that builds on the most popular deep reinforcement learning-based neural networks \cite{parisotto2015actor, mo2016personalizing, du2019improving}. These models usually focus on learning playing strategies for one game and transferring the learned state/action representations to similar tasks. They do obtain tangible performance gain when applied to similar scenarios (for example, similar Atari games \cite{rusu2016progressive}). \textcolor{black}{ Most of these models, however, rely on the transfer of world representation. The transfer of specific strategies, in particular on different scenarios, is usually neglected if the scenario representation changes considerably \cite{hua2021learning}}. Another common way to address transfer learning is through imitating an expert \cite{ho2016generative, fu2017learning}. In this type of learning, action/state transitions collected from an entity that knows how to solve the task are used to guide the learning of the agent. The agent usually derives a novel reward function to model the transitions from the expert. In imitation learning, an agent learns, thus, to perform the task similarly, and in most cases, better than the expert \cite{hester2017deep, hu2018deep}. Most of the scenarios where imitation learning was applied involve learning from humans \cite{stadie2017third, wang2019imitation, barde2020adversarial} or from a series of expert agents \cite{baram2017end, song2018multi, jeon2018bayesian} and although it allows for a direct individualized approach to learning, its applications to competitive games are limited \cite{swiechowski2020game}. Although these techniques can somehow shortcut the learning due to transferring basic and common representations, they usually fine-tune to new tasks by following the traditional extreme-generalization goal \cite{mittel2019visual}, which makes them unfitted for dealing with personalized strategies.



\subsection{The Competitive and Continual Learning Approach.} 

Although on a much smaller scale than the previously mentioned solutions, competitive reinforcement learning has been addressed in the past decades. The development of auxiliary experts by the DRON model, based on an opponent-specific deep Q Learning routine, was successful in learning how to counter the opponent's actions on simple games \cite{he2016opponent}. Also, observing the opponent's actions and using them to model a belief-based closed-world scenario, that is used to generate a Bayesian classifier was also explored \cite{yang2018towards}. Although this solution could model sophisticated strategies, going further from simplistic scenarios, it is extremely difficult to adapt towards novel opponents. Another way to explore competitive learning is through multi-agent interaction \cite{iqbal2019actor, li2019robust}, although, in most of these cases, the agents do not learn how to model the opponents' strategy, but treat the opponents as part of the environment dynamics, transforming the personalization problem into a generalization problem. The solutions can be seen as forms of continual reinforcement learning \cite{xu2018reinforced} approaches, that encapsulate techniques that allow a model to maintain a constant adaptation. In this regard, some models do this by prioritizing the selection of previous experiences \cite{rolnick2019experience, de2019continual} or adapting some sort of curriculum learning \cite{florensa2017reverse, shao2018starcraft}. In theory, continual learning, and thus competitive learning, is the ideal learning strategy to adapt a game-play style against specific opponents and still maintaining the general knowledge on how to play the game. What we see in reality, however, is a strong limitation on how to deal with competition on dynamic and multiplayer scenarios \cite{vamvoudakis2017game}, which we address with our proposed model.



\section{Opponent Specific Contrastive Reinforcement Learning}

Our proposed model, from here on referred as WINNE (the WINNer), is a modular architecture for individualized competitive behavior learning that can be used as an auxiliary reward estimator for any policy model. It has as a goal to specify an updated action that focuses on countering a possible action from an specific opponent. It relies on three basic requirements: \textbf{1) I know how to play the game; 2) I know how my opponent plays the game; 3) I know how to mitigate my opponent's actions.}

\begin{figure}
\centering
\includegraphics[width=1\columnwidth]{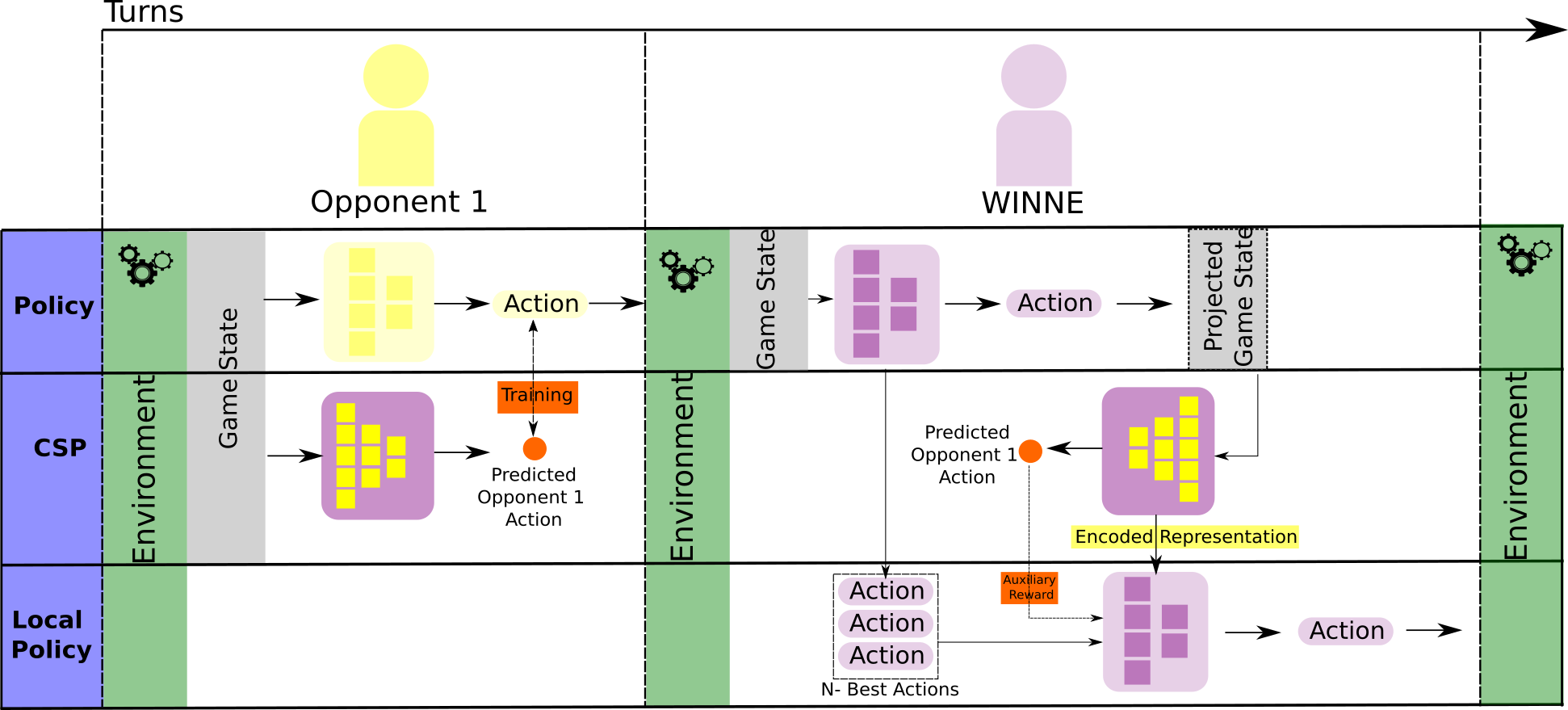}
\caption{\textcolor{black}{ The WINNE model. During an opponent round, it collects pairs of game states and actions to learn predicted actions for that specific opponent. When WINNE makes an action, it uses a global policy to determine an initial action, project the resulting game state when this action is selected, and predicts the opponent response using a Contrastive Strategy Prediction (CSP) layer. Using the entangled representations of the CSP and a set of possible actions coming from the global policy, a local policy network refines the final action WINNE will take by minimizing the chances of the predicted opponent response being chosen.}}
\label{fig:WINNE}
\end{figure}

The first requirement is addressed by a \textbf{global policy neural network} that focuses on learning a general strategy for the game. To achieve the second requirement, we introduce a \textbf{Contrastive Strategy Prediction (CSP) network} \textcolor{black}{that learns to predict opponent's actions}. The third requirement is addressed by a \textbf{local policy neural network}, that maps the representation learned by the CSP with a set of best possible actions coming from the global policy network.  \textcolor{black}{ WINNE integrates all these networks into an unified model, described in details in Figure \ref{fig:WINNE}, that is trained following one single optimization rule in an online manner.}


\subsection{Global Policy Network}


\textcolor{black}{The global policy network provides general knowledge about how to play the game successfully, thus any successfully policy network is suitable to be used as a global policy. Its goal within WINNE is to provide an initial estimation of possible actions, given the game state, that will allow WINNE to win the game. }

\subsection{Disrupting Opponent Actions Through CSP}

By observing the opponents' actions, WINNE must refine the initial possible game actions given by the global policy. Its goal is to generate an action that can disrupt the opponent's strategy, increasing the chances of victory. To complement the global policy network, which aims at learning the general game strategy, we rely on a contrastive learning routine to represent the opponent's strategies. Contrastive Learning \cite{chen2020simple} was presented recently as a self-supervised learning scheme that learns data representation through contrasting positive and negative samples. It was employed on predictive tasks \cite{oord2018representation}, supervised representation learning \cite{khosla2020supervised} and reinforcement learning \cite{laskin2020curl}, as it provides a fast representation learning method \cite{hsieh2020improving, banville2020uncovering}.

 Optimizing a supervised contrastive loss ($C_{loss}$) \cite{khosla2020supervised},  allows us to learn data representation that maximizes the ability of a model to identify different data points:

\begin{equation}
c_{i,j}= exp(z_{i} \cdot z_{j}/t)
c_{i,k} = exp(z_{i} \cdot z_{k}/t)
\end{equation}

\noindent where $i$ represents each data point on a training batch that shares the same labels with $j$, but has different labels than $k$, and $t$ is a temperature term defined before the training. This routine is applied for each update cycle in a training batch, and is used to calculate the contrastive supervised loss per game state:

\begin{equation}
L_i = \frac{-1}{N-1} \sum_{j=1}^N log \frac{c_{i,j}}{c_{i,k}}
\end{equation}


\noindent where $N$ is the total number of game state sets on that specific batch. The final loss to train CSP can be defined as the summation of all the game states on that batch:

\begin{equation}
L = \sum_{i=1}^N L_i
\end{equation}

\textcolor{black}{Our CSP network is trained on the opponent's rounds. It receives the observed game state as input and tries to map it with the taken actions. This layer is composed of a series of fully-connected Gated Recurrent Units (GRUs) \cite{chung2014empirical}, that allow the CSP to map the temporal dynamics between successive actions taken by an opponent. These layers act as an encoder and map the subsequent observed states with a taken action, represented by an output layer. The encoder and output layer are optimized using the contrastive loss $C_{loss}$, where we combine pairs of positive ($z_{i}$, sequences of data belonging to the same player) and negative ($z{j}$, sequences from other players scrambled) samples. For every player, WINNE creates one CSP layer, so we guarantee that each CSP will learn to predict the strategy of a single player. During the WINNE round, the global policy is used to select an initial action. We then project this action into the environment and created a projected game state. The CSP is used to predict the opponent's action in this projected game state, and we use a softmax activation in the output layer to obtain the probability that this action will be selected. This probability is used to map the prediction, from WINNE's perspective, of the opponent taking that action.}


The goal of the Contrastive Strategy Predictor (CSP) is to learn a rich representation that will encapsulate: \textcolor{black}{1) the understanding of the opponent's behavior, and in consequence, the opponent's game-play strategy, through the entangled representation of the encoder and 2) how to counter it, using the probability of the predicted action}.  We achieve this by observing the game state and action pairs of each of the opponents WINNE is playing against, and optimizing the CSP on these data.



\subsection{Local Policy Networks}

The local policy network has as a goal to distinguish which actions will disrupt the strategy of an opponent. As such, it must be a policy network that can be updated online, similarly to the global policy network. We feed it, however, with the entangled representation that comes from the CSP, concatenated with the $n$-best possible actions (selected as actions with the highest Q-values) coming from the global policy network. \textcolor{black}{ The $n$-best actions are represented as individualized and normalized values for each possible action, represented by the position of the individual output nodes of the global policy network.} The value of $n$ indicates the exploration capability of the local policy. We train it with a reinforcement learning routine, \textcolor{black}{using the inverse of the predicted action probability, from the CSP, as an auxiliary reward}, added to the original local policy reward, obtained from the environment. The goal of optimizing the local policy network is to minimize the auxiliary reward while maximizing the actions towards the environment end goal. 

\subsection{The Individualized Competition Approach}

\textcolor{black}{To maximize WINNE's capability of dealing with individual agents, each of them with their game-play strategy, WINNE creates a new instance of the CSP and local policy for each opponent it plays against. That means it will be able to associate individual playing styles per agent with strategies to disrupt the opponent's strategy. Coupling each individual CSP and local policy with a global policy, allows this transfer learning approach to derive individualized and successful strategies faster, when compared to end-to-end learning models.
The training of WINNE happens in two moments: first, we update the CSP during the opponent's actions, by coupling its actions with the current game state. Second, during WINNE's round, the local and global policies are updated using their reinforcement learning scheme, with the local policy receiving the auxiliary reward from the CSP. }

\section{Experimental Setup}

The evaluation of our model is based on the three requirements on which WINNE was based, and for each of them we perform a series of experiments with two scenarios: A Pokemon duel scenario, where WINNE plays a match against another agent, and a Chef's Hat scenario, where WINNE plays a game against four other agents.

\subsection{PokEnv}

Pokemon became a very popular game that is very easy to understand, but difficult to master. In a Pokemon duel, two opponents, controlling up to six Pokemon, battle against each other using one Pokemon at a time. It is a turn-based game, and in each turn, a player can choose one of the four moves a Pokemon has, or choose to switch an active Pokemon with another one in the team. The complexity of the game comes in terms of the combinatorial challenge to create a winning strategy. Up to today, there are 890 different pokemon and a total of 826 moves to choose from. Also, each pokemon has one or two of 16 different elemental types, which defines their weaknesses and strengths against each other. The same goes for each of the moves. The full set of rules can be found in the official competitive rule book \footnote{https://assets.pokemon.com//assets/cms2/pdf/play-pokemon/rules/play-pokemon-vg-rules-formats-and-penalty-guidelines-10232020-en.pdf}. 

In recent years, the complexity of the game raised the attention of the reinforcement learning community, and many implementations of simulated environments \cite{huang2019self,pagalyte2020go} appeared. These implementations and investigations, however, do not include the entire game mechanics or focus on a simplified version of the challenge, such as the choice of move selections based on type advantage \cite{simoes2020competitive}. In our experiments, we want to evaluate the capability of WINNE to learn how to model and counter a specific player's strategy within this hyper-complex environment. In this regard, we focus on the complexity of choosing the best actions, using the entire game mechanics. 

We use the PokEnv implementation for training artificial agents \cite{poke_env_2020_11_14} in all of our experiments. It uses the popular Pokemon Showdown simulator\footnote{https://pokemonshowdown.com/} to simulate the battles. The simulator reproduces all the current mechanics of the game and allows us to simulate battles using a random choice of Pokemon and moves. \textcolor{black}{ Figure \ref{fig:scenarios} illustrates the PokEnv environment.}

The PokEnv game state, used by the global policy network, is composed of 22 values, that represent the amount of Pokemon still in the game, their hit-points, and the power of the available moves. As the CSP does not have any knowledge about the opponent's private information, the CSP games state is composed of 10 values, including the Pokemon which are still in game and their hit-points. There are a total of 6 actions, that represent all the possible moves and Pokemon switch options. Agents are optimized using a standard reward function, defined by the authors of the environment, which gives a maximum reward if an opponent wins the match, but gives small boosts for defeating and performing attacks to the opponent's Pokemon.

\subsection{Chef's Hat GYM}

The four players Chef's Hat competitive card game \cite{barros2020s} is our second experimental scenario. At the beginning of the game, each player receives a set of cards, that represent different kitchen ingredients. The goal of the game is to be the first one to discard all the cards in your hand. The game is based on turns, and on each turn, a player can do a discard action or a pass action. For every match played, the players gain points based on their finishing position, with the winner gaining 3 points. A full game consists of several matches until one of the players reaches 15 points.

\begin{figure}[htbp]
\centering
\includegraphics[width=1\columnwidth]{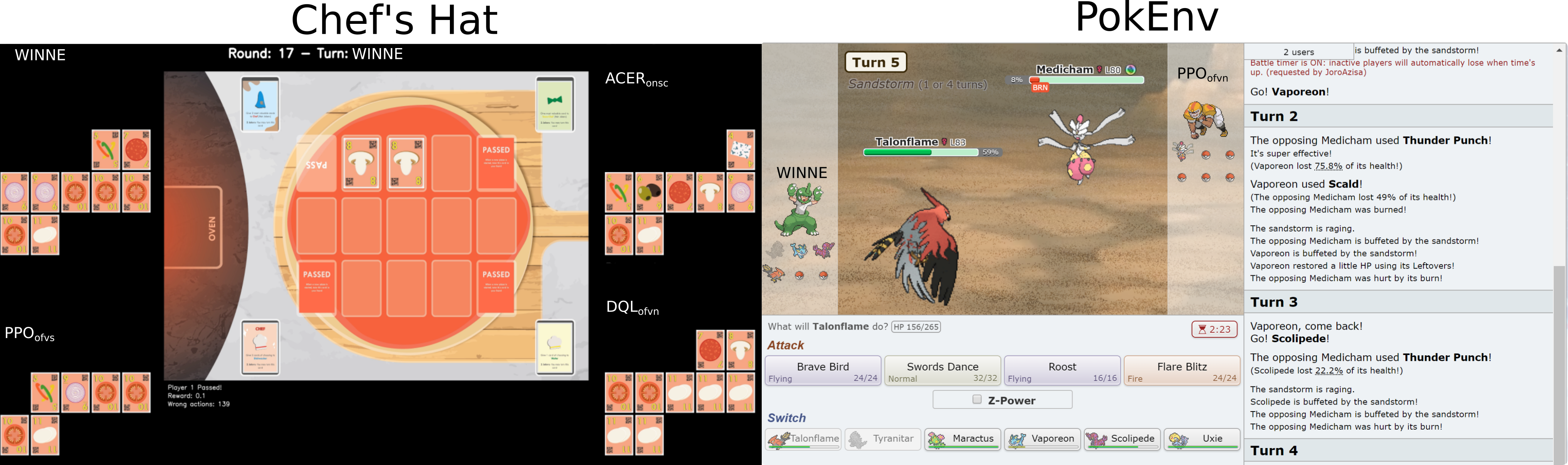}
\caption{Illustration of the scenarios we use to train and evaluate WINNE: Chef's Hat and PokEnv. }
\label{fig:scenarios}
\end{figure}

The game follows a series of simple rules for discarding cards, and the unique challenge is to overcome the other three players at the same time. Recent findings on training reinforcement learning agents on Chef's Hat \cite{barros2020learning} indicate that different algorithms enforce different game-play strategies with similar performance on the game.  \textbf{ In our experiments, we focus on evaluating how WINNE can deal with the individual strategies from the opponents when playing against different agents at the same time.}. \textcolor{black}{ Figure \ref{fig:scenarios} illustrates the PokEnv environment.}

The GYM implementation of Chef's Hat \footnote{https://github.com/pablovin/ChefsHatGYM} allows us to train artificial agents to play the game. The game state is composed of 28 values, which contain the total cards at hand and the board; the CSP game state is composed of 11 values, representing the cards on the board and an action selected by the opponent; and one action is represented by one-hot encoder of 200 possible actions. We follow the author's suggestions and use a global reward scheme that only gives a full reward once the player performs an action that leads to it winning the game.

\subsection{Implementation of WINNE and the Opponents}

As each scenario has its particularities, conceptually and technically, we implement two versions of WINNE, one for each scenario, and a series of opponents.

\subsubsection{WINNE implementation}

We performed a hyper-parameter tuning experiment to obtain the best global and local policies for each scenario. The same investigation served to define the opponents implementation, and all details are reported in our Appendix Section 1. The same for the CSP implementation, which is described in our Appendix Section 2.

\begin{figure}
\centering
\includegraphics[width=1\columnwidth]{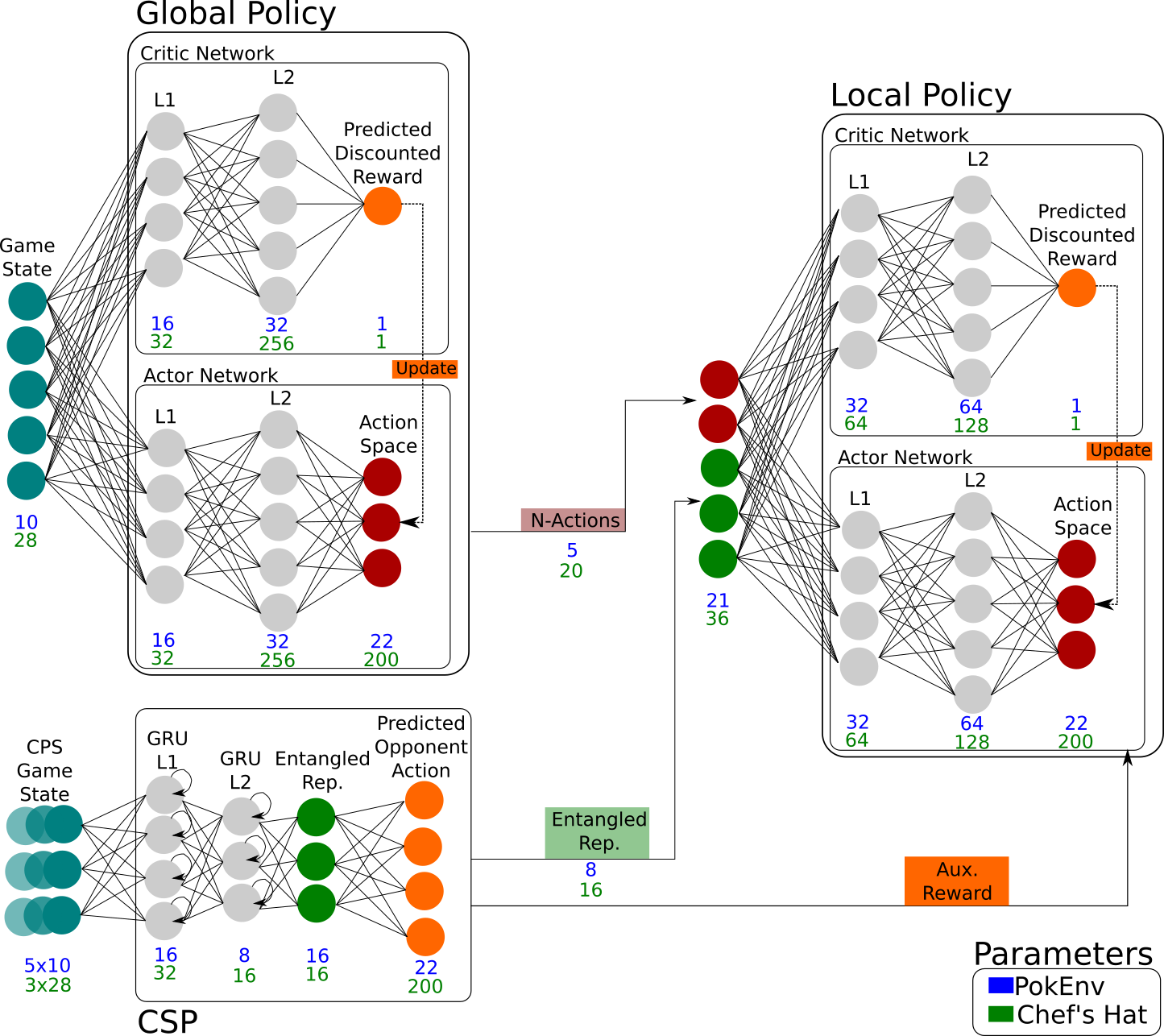}
\caption{Detailed implementation, with all the layers, connections and number of units of WINNE in our two scenarios: PokEnv and Chef's Hat, with implementation parameters in blue and green respectively. }
\label{fig:winneImplementation}
\end{figure}

\textcolor{black}{ The global and local policy networks are implemented as an actor-critic network composed of fully connected dense layers, trained with Proximal Policy Optimization (PPO).} The CSP network implements a GRU-based encoder, and all are described in detail in Figure \ref{fig:winneImplementation}.

\subsubsection{Opponents implementation}

To assess WINNE's capability of learning individualized competitiveness against different opponents, we implement four agents and six types of learning agents. The naive agents represent a unique strategy, or the lack of it, and do not learn. For the PokEnv scenario, the first naive agent performs random actions, while the second one only chooses moves that cause maximum damage to the opponent. On Chef's Hat, we also implement an agent that only provides random actions and one that discards always the maximum cards allowed.

The first four types of learning agents are implemented based on Deep Q-Learning (DQL) \cite{mnih2013playing}, Proximal Policy Optimization (PPO) \cite{schulman2017proximal}, Actor-Critic with Experience Replay (ACER) \cite{wang2016sample}, and Generative Adversarial Imitation Learning (GAIL) \cite{ho2016generative}. \textcolor{black}{ As GAIL requires expert demonstrations, we run 1000 games played by random agents and select the winning observations as guiding for the learning. In our experiments, this resulted in a competitive GAIL agent.}

As there are different manners to train these agents, and it would not be fair to compare a personalized approach with solutions that are not focused on learning personalization, we provide versions of these agents which are trained offline (playing against naive agents (\textbf{ofn}) and in self-play strategy (\textbf{ofn})) and online (learning from the scratch (\textbf{onsc}) and pre-trained against naive agents (\textbf{onpt})).

To provide a fair understanding of the impact of WINNE on competitive learning, we also implement and evaluate two competitive-based solutions: DRON (Deep Reinforcement Opponent Network) \cite{he2016opponent} and the Bayes-ToMoP \cite{yang2018towards} models. Both these models continue learning while the game happens, and thus, are provided only as an online setting. 

\textcolor{black}{ In total, we have 18 learning agents (four learning settings for each of the four agent implementations, plus two competitive-based agents) and 2 naive agents for each scenario. }

Table \ref{tab:opponents} summarizes all the types of opponents and their descriptions.

\begin{table}[h!]
\center
\begin{tabular}{ |c| c|  c| c |} 
\hline
 \textbf{Name} & \textbf{Type}  & \textbf{Training} & \textbf{Strategy}  \\ \hline
 $N_r$ & Naive & - & Random\\
 $N_s$ & Naive &  - & Simple Strategy \\
 ofsp & Learning &  Offline & Self-play \\
 ofvn & Learning &  Offline & vs Naive \\
 onsc & Learning &  Online & Scratch \\
 onpt & Learning &  Online & Pre-trained\\ \hline
 DRON & Competitive & Online & - \\ 
 Bayes-ToMoP & Competitive & Online & -\\\hline
\end{tabular}
\caption{Types of opponent implementations. Each of the learning agents implements versions of DQL, PPO, ACER, and GAIL agents, totaling 16 different types of agents. The competitive agents follow the original authors´ implementation.}
\label{tab:opponents}
\end{table}

\subsection{The Tournament Scenario}

In order to provide a holistic experimental setup, without requiring an enormous set of combinatorial experiments, our main evaluation scenarios happen in the form of a tournament between WINNE and all its opponents.

Each tournament is composed of two brackets of hierarchical playing phases. In each phase, the agents will face each other, and the victorious ones will advance to the next phase and play against themselves, until one single agent wins the tournament. For PokEnv, each group is formed of a duel, and the winner advances to the next phase. In Chef's Hat, four players form a group, and the two best players advance to the next phase.

We implement one instance of each agent, to allow a fair behavioral comparison. To complement the game to allow the tournament setting, we instantiate 14 naive agents.


\subsection{Experiments and Metrics}

To evaluate requirement \textbf{1) I know how to play the game}, we run a \textbf{benchmark tournament} where we put all the agents to play against each other. In this phase, we are evaluating how the global policy network performs when playing the game against all other 31 agents. We run 10 tournaments in a row, allowing the continual learning models to update while playing. We repeat this scenario 100 times and calculate the average number of individual victories per tournament run. 

Requirement 2) focuses on assessing how well the CSP network can \textbf{learn strategies from each type of opponent}. In this regard, we run 10 individual games where WINNE plays only against instances of opponents of the same type. We measure the accuracy of predicting sequences of actions from the same opponent.

Our third experiment focus on assessing requirement 3). We will run the same tournament setup described in our first experiment, but using the full implementation of WINNE. In this experiment, we measure the individual performance of WINNE against each of the agents, in terms of individual victories divided by the number of games played, per tournament run. This allows us to understand the \textbf{impact of the local policy network on individual agent types}. 

\textcolor{black}{ We run a fourth and final experiment, to assess the capability of WINNE to retain knowledge when playing against the same agents after longer intervals. In this experiment, we run the full implementation of WINNE
and set up a 10 cycles game, wherein each cycle WINNE plays 10 games against each of the 18 learning opponents. This means that WINNE will play against one opponent type for 10 games, and play against this opponent again only after 170 games. We calculate the average performance of WINNE in terms of victories against the specific agents over the 10 cycles. This experiment aims to assess how well WINNE maintains the competitive knowledge against specific agent types, even after playing against different agents for longer periods. }

\section{Results}

\subsection{Experiment 1: How to play the game.}

\begin{figure}[htbp]
\centering
\includegraphics[width=1\columnwidth]{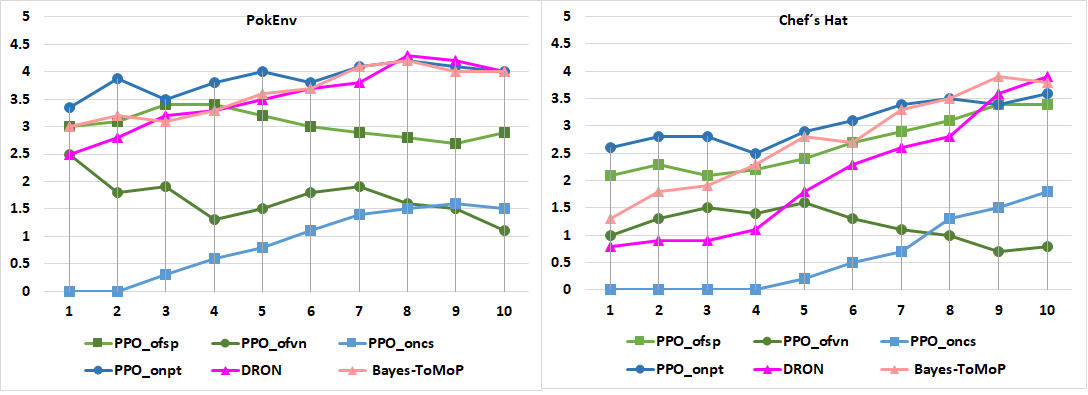}
\caption{Averaged number of victories for each of the 10 tournaments, over the 100 runs, per agent type.}
\label{fig:result1}
\end{figure}

Our first experimental results, summarized in Figure \ref{fig:result1}, give us an insight into the different performances of each of the agents' implementations. As expected, the naive agents do not present a good performance, and for brevity, we report their performance on the supplementary material Section 3. In both scenarios, offline agents start with a good performance and present a drop over time, impacted by the online agents. In terms of performance, the online agents that are pre-trained present good overall average results at the end of the 10 tournaments, which is expected as they continue to adapt towards the scenario, differently from the offline agents. The competitive agents, DRON and Bayes-ToMoP present a behavior similar to our online learning agents, but achieving the best overall results at the end of the 10 tournaments.


\subsection{Experiment 2: How to predict an opponent's action.}

\begin{figure}[htbp]
\centering
\includegraphics[width=1\columnwidth]{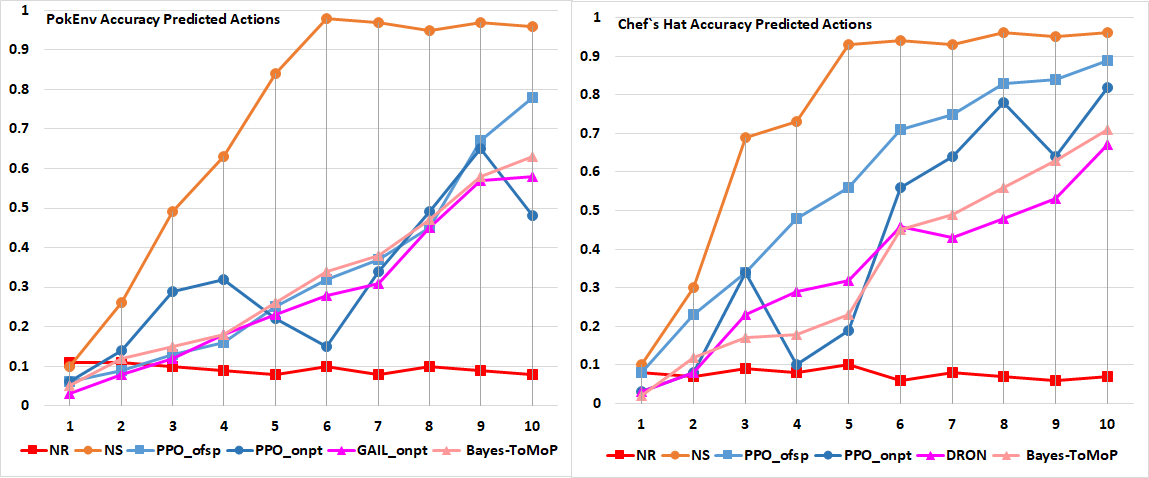}
\caption{Average accuracy between the CSP's predictions of a sequence of observations belonging to a specific agent, over 10 tournaments and 100 runs. }
\label{fig:results2}
\end{figure}

Our CSP network relies on an opponent using a specific strategy, or a set of strategies, in order to learn it. In figure \ref{fig:results2}, we plot the accuracy of the predictions over 10 games for the actions performed by naive agents, an example of the best offline ($PPO_{ofsp}$) and online ($PPO_{onpt}$) agents, and the DRON and Bayes-ToMoP agents. We report the same analysis for all the agents in our supplementary material Section 4. We observe that the naive agent with the simple strategy ($N_S$) is the easiest to predict, achieving an accuracy over 90\% after playing 6 games against WINNE on the PokEnv environment, and even earlier with 5 games on Chef's Hat. The random naive agent ($N_r$), however, does not have a specific strategy to play the game, thus, WINNE cannot predict its next moves. In both scenarios, WINNE can predict the offline agent's actions over time, each game providing a better prediction than the previous one. The online agent adapts its playing style over time, and WINNE needs more examples to predict its actions with higher accuracy. The same occurs with the competitive-based agents.


\subsection{Experiment 3: Adapting towards individual opponents.}

This experiment measures the capability of WINNE of adapting to each of the opponents and we plot the percentage of victories WINNE has over each opponent over the 10 tournaments, averaged for 100 runs, in Figure \ref{fig:result3}. Again, we display here only the percentage against the two naive agents, and the best offline ($PPO_{ofsp}$), online ($PPO_{onpt}$), and both DRON and Bayes-ToMoP agents, but all the results are available in our supplementary material Section 5. In accordance with our previous experiments, we observe that WINNE struggles to play against the naive random agent ($N_r$), as it does not learn any counter-strategy to beat it. Against the other agents, in particular against the naive agent with a simple strategy ($N_s$), WINNE thrives. The competitive agents display an interesting behavior, as they also adapt towards WINNE. At the end of 8 games, WINNE can overcome their adaptation and increases the number of victories. 

\begin{figure}[htbp]
\centering
\includegraphics[width=1\columnwidth]{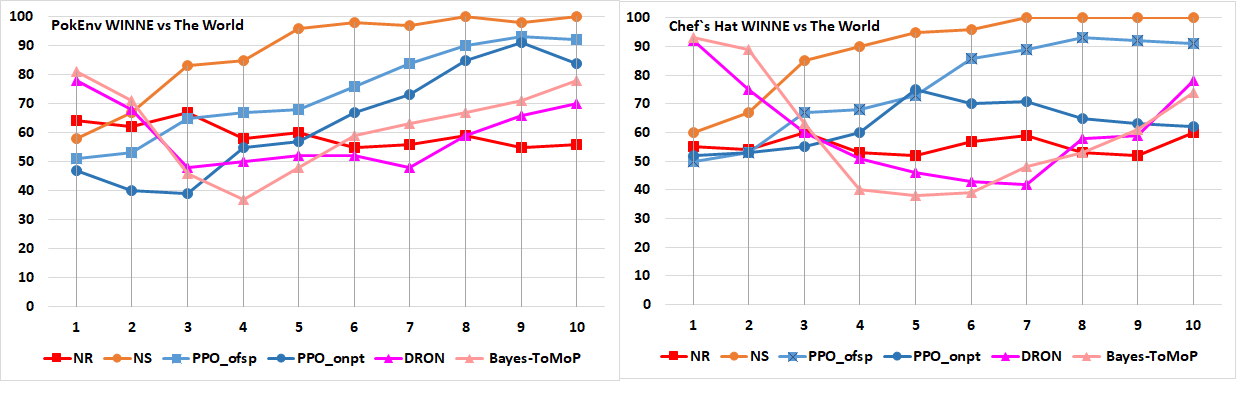}
\caption{Averaged percentage of victories of WINNE over specific agents, in 10 tournaments and 100 runs.}
\label{fig:result3}
\end{figure}

\subsection{Experiment 4: Retaining Knowledge Over a Longer Period of Time.}

\begin{figure}[htbp]
\centering
\includegraphics[width=1\columnwidth]{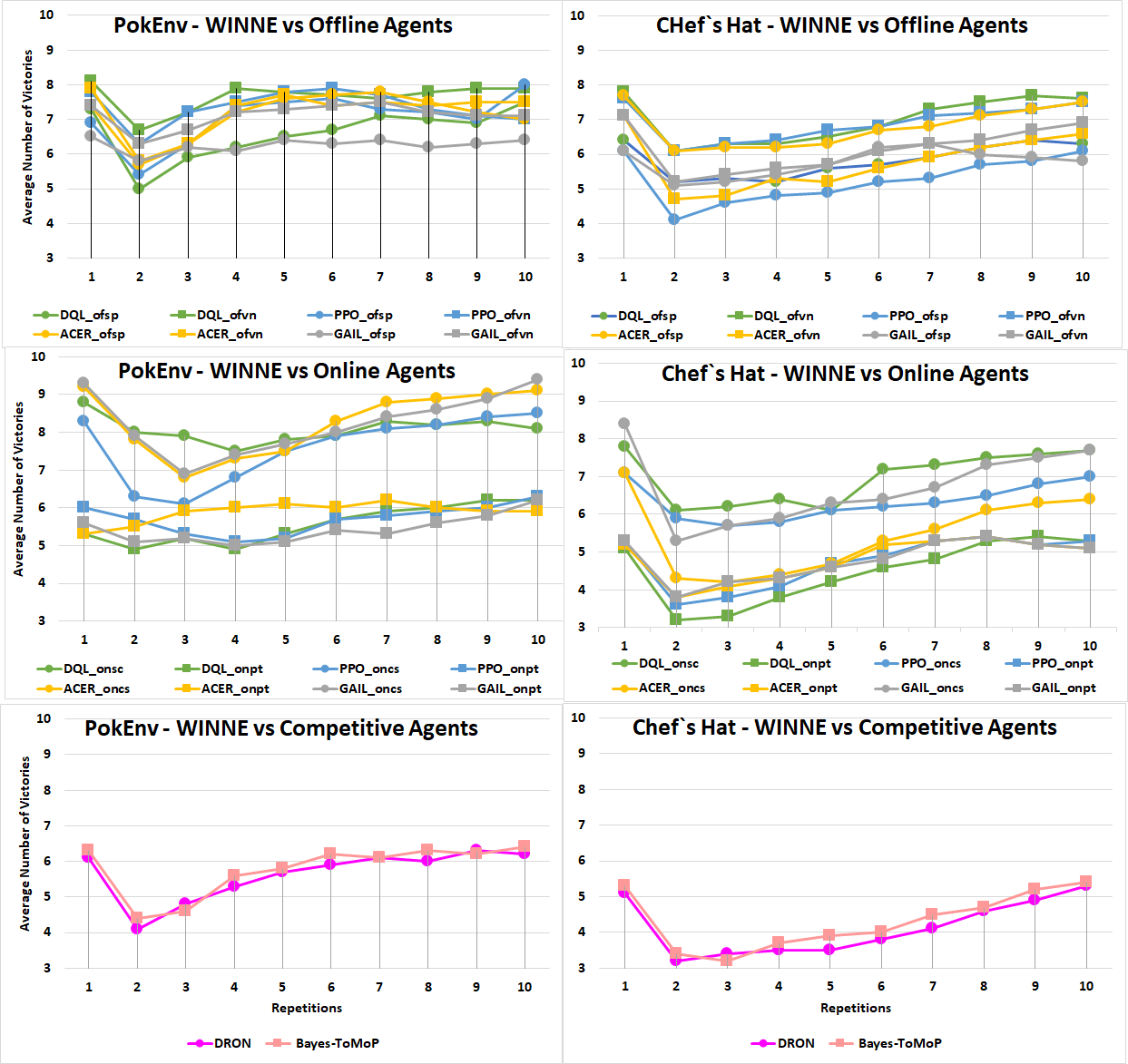}
\caption{Averaged number of victories of WINNE over specific agent types for 10 cycles. In each cycle, WINNE plays 10 times against each of the learning agents type.}
\label{fig:result4}
\end{figure}

\textcolor{black}{  In our fourth experiment, reported in Figure \ref{fig:result4}, we observe the performance of WINNE playing against the same type of agents after an interval of 170 games. Against all 18 types of agents, WINNE presents the same behavior, dropping the performance in the second game, when compared to the first. In particular, when playing against online and competitive opponents, the performance drop is higher. After the second game, however, WINNE seems to recover rather quickly, and the performance seems to increase at a steady pace. In some cases, in particular against the naive online agents on the PokEnv, the performance after the 10 played cycles increases above the initial performance.}

\section{Discussions and Future Work}

Our experiments display the capability of WINNE to deal with the opponents in both the evaluated scenarios. In general terms, by leveraging on the pre-trained global policy, WINNE performs well from the beginning, and improves its performance every time it plays against the agents. In this section, we discuss how WINNE affects the general behavior of each agent within the tournament scenario for each environment. 

\subsection{PokEnv and Chef`s Hat Particularities}

The PokEnv scenario showed to be the most challenging to be learned online, in particular for agents without any prior knowledge, as shown by the evolution of the continual learning and competitive agents. WINNE, thus, presents a smaller performance improvement when playing this scenario in terms of individual victories, as exhibited in Figure \ref{fig:discussion1}. Once again for brevity here we display only the results of all the PPO and competitive-related agents. The entire report is available in our supplementary material Section 6. The decrease in performance of the competitive-based agents happens because, as indicated by our third experiment, WINNE can predict these agents' moves, even if they keep adapting. In general, the good performance of WINNE prevents the opponents from learning a strategy to beat it.

\begin{figure}[htbp]
\centering
\includegraphics[width=1\columnwidth]{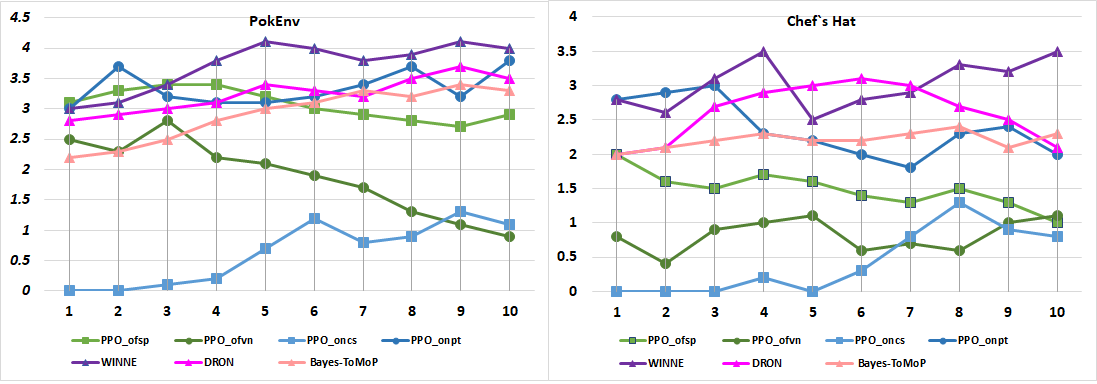}
\caption{Performance, per tournament round, averaged over 100 runs of the games with and without WINNE on the PokEnv and Chef's Hat environment.}
\label{fig:discussion1}
\end{figure}

On Chef's Hat, WINNE's behavior is similar to that observed with PokEnv, but with an even greater impact on the opponent's performance. As in Chef's Hat each game is played against three different opponents at the time, WINNE can learn how to understand their game strategy in a single game, having much more information to counter them, when compared to the competitive-based agents that only focus on learning the strategy of one agent per time.

\subsection{The Impact of Long-term learning}

\textcolor{black}{ Our last experiment demonstrates the impact of long-term learning on WINNE. The performance drop indicates that the contribution of the CSP and the local policy for individualized strategy learning has an impact on the global policy learning, although the general performance of WINNE is still high. However, the presence of the individualized CSP and local policy shows a quick recovery in performance in all scenarios, including the more challenging competitive agents: DROn and Bayes-ToMop. Therefore, the combination of joint optimization of the three modules has a positive impact on WINNE`s behavior as a long-term learning agent, although it can struggle on the first interactions. 
Our results show that the balance between generalized and personalized strategy gives WINNE a tool to escape the catastrophic forgetting problem, common on continual and transfer learning models. Optimizing the performance drop after the first game might be an interesting future development, and recent studies on extended memory replay \cite{atkinson2021pseudo} might help to address this problem.}

\subsection{The Lack fo Strategy Problem}

The entire conception of WINNE is based on one assumption: my opponent knows how to play the game. That is how the CSP network can learn how to embed opponent observations, and that is why WINNE presents a strategy disruption early on most of the games it played. When playing against opponents that do not have a strategy, WINNE cannot adapt as there is no strategy to be learned. This is expected, based on how the model was conceived, and this situation usually happens only in artificially created situations, as humans tend to follow a strategy \cite{west2006cognitive}.

\subsection{What is next?}

Although tailored and applied to competitive reinforcement learning, we believe that the CSP, in particular, can be used as a learning strategy for different scenarios. The contrastive optimization can be used as an auxiliary guide for reinforcement learning. We envision its use in identifying contextual information, by contrasting different grounded characteristics from the environment itself, such as dynamic changes or environmental characteristics. 

Updating our model to also learn high-dimensional state representation might help to adapt it too complex video games scenarios. Also, on the Chef's Hat game, we believe that adopting the prediction network to deal with discounted actions to take into consideration the entire game dynamics merging the individual strategies, could improve even further the models' performance.

\section{Conclusion}

In this paper, we introduced a novel contrastive reinforcement learning model that allows an agent, implemented as WINNE, to adapt its decision making towards specific opponents when playing a pokemon duel and the Chef's Hat competitive card game.

Our model relies on three principles: prior-knowledge about the task, represented by global policy network; learning opponent's strategic behavior, done by a contrastive predictive neural network; and changing its behavior to counter the opponent's actions, using individual local policy networks per adversary. The entire model showed to improve WINNE's capability of playing both games, even when facing adversaries that present a continual  and competitive learning behavior. WINNE was able not only to adapt towards many different types of opponents, but also to learn how to disrupt them. Our experiments show that specific types of opponent's loose performance quickly when playing consecutive games against WINNE.

\section{Acknowledgment}

A.S. is supported by a Starting Grant from the European Research Council (ERC) under the European Union's Horizon 2020 research and innovation programme. G.A. No 804388, wHiSPER. PB greets in particular the real WINNE for the years of support and friendship. We finally did a player that plays Pokemon better than you.

{\small

\bibliography{biblio}

\begin{thebibliography}{10}
\expandafter\ifx\csname url\endcsname\relax
  \def\url#1{\texttt{#1}}\fi
\expandafter\ifx\csname urlprefix\endcsname\relax\def\urlprefix{URL }\fi
\expandafter\ifx\csname href\endcsname\relax
  \def\href#1#2{#2} \def\path#1{#1}\fi

\bibitem{lanctot2017unified}
M.~Lanctot, V.~Zambaldi, A.~Gruslys, A.~Lazaridou, K.~Tuyls, J.~P{\'e}rolat,
  D.~Silver, T.~Graepel, A unified game-theoretic approach to multiagent
  reinforcement learning, in: Advances in neural information processing
  systems, 2017, pp. 4190--4203.

\bibitem{silver2018general}
D.~Silver, T.~Hubert, J.~Schrittwieser, I.~Antonoglou, M.~Lai, A.~Guez,
  M.~Lanctot, L.~Sifre, D.~Kumaran, T.~Graepel, et~al., A general reinforcement
  learning algorithm that masters chess, shogi, and go through self-play,
  Science 362~(6419) (2018) 1140--1144.

\bibitem{torrado2018deep}
R.~R. Torrado, P.~Bontrager, J.~Togelius, J.~Liu, D.~Perez-Liebana, Deep
  reinforcement learning for general video game ai, in: 2018 IEEE Conference on
  Computational Intelligence and Games (CIG), IEEE, 2018, pp. 1--8.

\bibitem{chen2020simple}
T.~Chen, S.~Kornblith, M.~Norouzi, G.~Hinton, A simple framework for
  contrastive learning of visual representations, in: International conference
  on machine learning, PMLR, 2020, pp. 1597--1607.

\bibitem{poke_env_2020_11_14}
H.~Sahovic, \href{https://github.com/hsahovic/poke-env}{Poke-env: pokemon ai in
  python}.
\newline\urlprefix\url{https://github.com/hsahovic/poke-env}

\bibitem{barros2020learning}
P.~Barros, A.~Tanevska, A.~Sciutti, Learning from learners: Adapting
  reinforcement learning agents to be competitive in a card game, arXiv
  preprint arXiv:2004.04000.

\bibitem{mnih2013playing}
V.~Mnih, K.~Kavukcuoglu, D.~Silver, A.~Graves, I.~Antonoglou, D.~Wierstra,
  M.~Riedmiller, Playing atari with deep reinforcement learning, arXiv preprint
  arXiv:1312.5602.

\bibitem{schulman2017proximal}
J.~Schulman, F.~Wolski, P.~Dhariwal, A.~Radford, O.~Klimov, Proximal policy
  optimization algorithms, arXiv preprint arXiv:1707.06347.

\bibitem{wang2016sample}
Z.~Wang, V.~Bapst, N.~Heess, V.~Mnih, R.~Munos, K.~Kavukcuoglu, N.~de~Freitas,
  Sample efficient actor-critic with experience replay, arXiv preprint
  arXiv:1611.01224.

\bibitem{ho2016generative}
J.~Ho, S.~Ermon, Generative adversarial imitation learning, in: Advances in
  neural information processing systems, 2016, pp. 4565--4573.

\bibitem{he2016opponent}
H.~He, J.~Boyd-Graber, K.~Kwok, H.~Daum{\'e}~III, Opponent modeling in deep
  reinforcement learning, in: International conference on machine learning,
  PMLR, 2016, pp. 1804--1813.

\bibitem{yang2018towards}
T.~Yang, Z.~Meng, J.~Hao, C.~Zhang, Y.~Zheng, Z.~Zheng, Towards efficient
  detection and optimal response against sophisticated opponents, arXiv
  preprint arXiv:1809.04240.

\bibitem{parisotto2015actor}
E.~Parisotto, J.~L. Ba, R.~Salakhutdinov, Actor-mimic: Deep multitask and
  transfer reinforcement learning, arXiv preprint arXiv:1511.06342.

\bibitem{mo2016personalizing}
K.~Mo, S.~Li, Y.~Zhang, J.~Li, Q.~Yang, Personalizing a dialogue system with
  transfer reinforcement learning, arXiv preprint arXiv:1610.02891.

\bibitem{du2019improving}
Y.~Du, Improving deep reinforcement learning via transfer, in: Proceedings of
  the 18th International Conference on Autonomous Agents and MultiAgent
  Systems, 2019, pp. 2405--2407.

\bibitem{rusu2016progressive}
A.~A. Rusu, N.~C. Rabinowitz, G.~Desjardins, H.~Soyer, J.~Kirkpatrick,
  K.~Kavukcuoglu, R.~Pascanu, R.~Hadsell, Progressive neural networks, arXiv
  preprint arXiv:1606.04671.

\bibitem{hua2021learning}
J.~Hua, L.~Zeng, G.~Li, Z.~Ju, Learning for a robot: Deep reinforcement
  learning, imitation learning, transfer learning, Sensors 21~(4) (2021) 1278.

\bibitem{fu2017learning}
J.~Fu, K.~Luo, S.~Levine, Learning robust rewards with adversarial inverse
  reinforcement learning, arXiv preprint arXiv:1710.11248.

\bibitem{hester2017deep}
T.~Hester, M.~Vecerik, O.~Pietquin, M.~Lanctot, T.~Schaul, B.~Piot, D.~Horgan,
  J.~Quan, A.~Sendonaris, G.~Dulac-Arnold, et~al., Deep q-learning from
  demonstrations, arXiv preprint arXiv:1704.03732.

\bibitem{hu2018deep}
Z.~Hu, Z.~Yang, R.~R. Salakhutdinov, L.~Qin, X.~Liang, H.~Dong, E.~P. Xing,
  Deep generative models with learnable knowledge constraints, in: Advances in
  Neural Information Processing Systems, 2018, pp. 10501--10512.

\bibitem{stadie2017third}
B.~C. Stadie, P.~Abbeel, I.~Sutskever, Third-person imitation learning, arXiv
  preprint arXiv:1703.01703.

\bibitem{wang2019imitation}
B.~Wang, E.~Adeli, H.-k. Chiu, D.-A. Huang, J.~C. Niebles, Imitation learning
  for human pose prediction, in: Proceedings of the IEEE International
  Conference on Computer Vision, 2019, pp. 7124--7133.

\bibitem{barde2020adversarial}
P.~Barde, J.~Roy, W.~Jeon, J.~Pineau, C.~Pal, D.~Nowrouzezahrai, Adversarial
  soft advantage fitting: Imitation learning without policy optimization, arXiv
  preprint arXiv:2006.13258.

\bibitem{baram2017end}
N.~Baram, O.~Anschel, I.~Caspi, S.~Mannor, End-to-end differentiable
  adversarial imitation learning, in: International Conference on Machine
  Learning, 2017, pp. 390--399.

\bibitem{song2018multi}
J.~Song, H.~Ren, D.~Sadigh, S.~Ermon, Multi-agent generative adversarial
  imitation learning, in: Advances in neural information processing systems,
  2018, pp. 7461--7472.

\bibitem{jeon2018bayesian}
W.~Jeon, S.~Seo, K.-E. Kim, A bayesian approach to generative adversarial
  imitation learning, in: Advances in Neural Information Processing Systems,
  2018, pp. 7429--7439.

\bibitem{swiechowski2020game}
M.~{\'S}wiechowski, Game ai competitions: Motivation for the imitation
  game-playing competition, in: 2020 15th Conference on Computer Science and
  Information Systems (FedCSIS), IEEE, 2020, pp. 155--160.

\bibitem{mittel2019visual}
A.~Mittel, P.~Sowmya~Munukutla, Visual transfer between atari games using
  competitive reinforcement learning, in: Proceedings of the IEEE Conference on
  Computer Vision and Pattern Recognition Workshops, 2019, pp. 0--0.

\bibitem{iqbal2019actor}
S.~Iqbal, F.~Sha, Actor-attention-critic for multi-agent reinforcement
  learning, in: International Conference on Machine Learning, PMLR, 2019, pp.
  2961--2970.

\bibitem{li2019robust}
S.~Li, Y.~Wu, X.~Cui, H.~Dong, F.~Fang, S.~Russell, Robust multi-agent
  reinforcement learning via minimax deep deterministic policy gradient, in:
  Proceedings of the AAAI Conference on Artificial Intelligence, Vol.~33, 2019,
  pp. 4213--4220.

\bibitem{xu2018reinforced}
J.~Xu, Z.~Zhu, Reinforced continual learning, in: Advances in Neural
  Information Processing Systems, 2018, pp. 899--908.

\bibitem{rolnick2019experience}
D.~Rolnick, A.~Ahuja, J.~Schwarz, T.~Lillicrap, G.~Wayne, Experience replay for
  continual learning, in: Advances in Neural Information Processing Systems,
  2019, pp. 350--360.

\bibitem{de2019continual}
M.~De~Lange, R.~Aljundi, M.~Masana, S.~Parisot, X.~Jia, A.~Leonardis,
  G.~Slabaugh, T.~Tuytelaars, Continual learning: A comparative study on how to
  defy forgetting in classification tasks, arXiv preprint arXiv:1909.08383
  2~(6).

\bibitem{florensa2017reverse}
C.~Florensa, D.~Held, M.~Wulfmeier, M.~Zhang, P.~Abbeel, Reverse curriculum
  generation for reinforcement learning, arXiv preprint arXiv:1707.05300.

\bibitem{shao2018starcraft}
K.~Shao, Y.~Zhu, D.~Zhao, Starcraft micromanagement with reinforcement learning
  and curriculum transfer learning, IEEE Transactions on Emerging Topics in
  Computational Intelligence 3~(1) (2018) 73--84.

\bibitem{vamvoudakis2017game}
K.~G. Vamvoudakis, H.~Modares, B.~Kiumarsi, F.~L. Lewis, Game theory-based
  control system algorithms with real-time reinforcement learning: How to solve
  multiplayer games online, IEEE Control Systems Magazine 37~(1) (2017) 33--52.

\bibitem{oord2018representation}
A.~v.~d. Oord, Y.~Li, O.~Vinyals, Representation learning with contrastive
  predictive coding, arXiv preprint arXiv:1807.03748.

\bibitem{khosla2020supervised}
P.~Khosla, P.~Teterwak, C.~Wang, A.~Sarna, Y.~Tian, P.~Isola, A.~Maschinot,
  C.~Liu, D.~Krishnan, Supervised contrastive learning, arXiv preprint
  arXiv:2004.11362.

\bibitem{laskin2020curl}
M.~Laskin, A.~Srinivas, P.~Abbeel, Curl: Contrastive unsupervised
  representations for reinforcement learning, in: International Conference on
  Machine Learning, PMLR, 2020, pp. 5639--5650.

\bibitem{hsieh2020improving}
T.-A. Hsieh, C.~Yu, S.-W. Fu, X.~Lu, Y.~Tsao, Improving perceptual quality by
  phone-fortified perceptual loss for speech enhancement, arXiv preprint
  arXiv:2010.15174.

\bibitem{banville2020uncovering}
H.~Banville, O.~Chehab, A.~Hyv{\"a}rinen, D.-A. Engemann, A.~Gramfort,
  Uncovering the structure of clinical eeg signals with self-supervised
  learning, arXiv preprint arXiv:2007.16104.

\bibitem{chung2014empirical}
J.~Chung, C.~Gulcehre, K.~Cho, Y.~Bengio, Empirical evaluation of gated
  recurrent neural networks on sequence modeling, arXiv preprint
  arXiv:1412.3555.

\bibitem{huang2019self}
D.~Huang, S.~Lee, A self-play policy optimization approach to battling
  pok{\'e}mon, in: 2019 IEEE Conference on Games (CoG), IEEE, 2019, pp. 1--4.

\bibitem{pagalyte2020go}
E.~Pagalyte, M.~Mancini, L.~Climent, Go with the flow: Reinforcement learning
  in turn-based battle video games, in: Proceedings of the 20th ACM
  International Conference on Intelligent Virtual Agents, 2020, pp. 1--8.

\bibitem{simoes2020competitive}
D.~Sim{\~o}es, S.~Reis, N.~Lau, L.~P. Reis, Competitive deep reinforcement
  learning over a pok{\'e}mon battling simulator, in: 2020 IEEE International
  Conference on Autonomous Robot Systems and Competitions (ICARSC), IEEE, 2020,
  pp. 40--45.

\bibitem{barros2020s}
P.~Barros, A.~Sciutti, I.~M. Hootsmans, L.~M. Opheij, R.~H. Toebosch,
  E.~Barakova, It's food fight! introducing the chef's hat card game for
  affective-aware hri, arXiv preprint arXiv:2002.11458.

\bibitem{atkinson2021pseudo}
C.~Atkinson, B.~McCane, L.~Szymanski, A.~Robins, Pseudo-rehearsal: Achieving
  deep reinforcement learning without catastrophic forgetting, Neurocomputing
  428 (2021) 291--307.

\bibitem{west2006cognitive}
R.~L. West, C.~Lebiere, D.~J. Bothell, Cognitive architectures, game playing,
  and human evolution, Cognition and multi-agent interaction: From cognitive
  modeling to social simulation (2006) 103--123.

\bibitem{bergstra2015hyperopt}
J.~Bergstra, B.~Komer, C.~Eliasmith, D.~Yamins, D.~D. Cox, Hyperopt: a python
  library for model selection and hyperparameter optimization, Computational
  Science \& Discovery 8~(1) (2015) 014008.

\end{thebibliography}
}

\appendix

\section{Appendix}

\subsection{Agent's Implementation}

For our experiments, we implement four different reinforcement learning agents ( Deep Q-Learning (DQL) \cite{mnih2013playing}, Proximal Policy Optimization (PPO) \cite{schulman2017proximal}, Actor-Critic with Experience Replay (ACER) \cite{wang2016sample}, and Generative Adversarial Imitatio Learning (GAIL) \cite{ho2016generative} ), following an offline pre-training and a continual online learning strategy. In total, we developed 16 agents. In this section, we display the hyper-parameter tuning decisions and the final architecture of each of the agents.

All of our agents were fine-tuned using the Hyperopt library \cite{bergstra2015hyperopt}. We run 1000 evaluations per agent, and use the one with the best performance for all experiments. Performance is measured on the number of combined victories when playing 1000 games against 500 random agents, and 500 simple strategy agents.

The GAIL agent learns based on expert demonstrations. To generate the demonstrations, we run 1000 games of two random agents and select the victorious games as experts.

\textbf{Parameters Search Space}

We run one exploration search per scenario, each of them with the search spaces defined in Table \ref{tab:searchSpaceAll} for all the agents.

\begin{table}[h!]
\center
\begin{tabular}{ |c|c|} 
\hline
 \textbf{Parameter} & \textbf{Search Space} \\ \hline
 \multicolumn{2}{|c|}{\textbf{DQL}}\\\hline
 Number of Layers & [1,2,3]\\
 Units Per Layer & [16, 32, 64, 128, 256, 512, 1024] \\
 Gamma & [0.5; 0.99] \\
 Learning Rate & [0.0005; 0.9]\\
 Double Q Learning & [True, False]\\
 Target Net. Updates &  [10, 50, 100, 250, 500, 1000]\\
 Prioritize Replay &  [True, False]\\
 \hline
  \multicolumn{2}{|c|}{\textbf{PPO}}\\\hline
  Number of Layers & [1,2,3]\\
 Units Per Layer & [16, 32, 64, 128, 256, 512, 1024] \\
 Gamma & [0.5; 0.99] \\
 Learning Rate & [0.0005; 0.9]\\
 Entropy Coeff. & [0.001; 0.9]\\
 Value Func. Coeff. &  [0.1; 0.9]\\
  \hline
  \multicolumn{2}{|c|}{\textbf{ACER}}\\\hline
   Number of Layers & [1,2,3]\\
 Units Per Layer & [16, 32, 64, 128, 256, 512, 1024] \\
 Gamma & [0.5; 0.99] \\
 Learning Rate & [0.0005; 0.5]\\
 Entropy Coefficient & [0.001; 0.9]\\
 buffer size &  [100, 500, 1000, 3000, 5000, 10000]\\
   \hline
  \multicolumn{2}{|c|}{\textbf{GAIL}}\\\hline
  Number of Layers & [1,2,3]\\
   Units Per Layer & [16, 32, 64, 128, 256, 512, 1024] \\
 Gamma & [0.5; 0.99] \\
 Learning Rate & [0.0005; 0.9]\\
 Entropy Coeff. & [0.001; 0.9]\\
 MAx KL Loss &  [0.0001 ; 0.9]\\\hline
\end{tabular}
\caption{Search space used to optimize all of our agents in both scenarios.}
\label{tab:searchSpaceAll}
\end{table}

 

 

 

\textbf{Final Architecture Opponents}

Table \ref{tab:finalArchitecturesAll} displays the final architecture used for both scenario for all of our agents.

\begin{table}[h!]
\center
\begin{tabular}{ |c|c|c|} 
\hline
 \textbf{Parameter} & \textbf{PokEnv} & \textbf{Chef's Hat} \\ \hline
   \hline
  \multicolumn{3}{|c|}{\textbf{DQL}}\\\hline
 Number of Layers & 2 & 2\\
 Units Per Layer & 16, 256& 32, 256\\
 Gamma & 0.95 & 0.98 \\
 Learning Rate & 0.03 &  0.004\\
 Double Q Learning & True & True\\
 Target Network update & 500 & 500 \\
 Prioritize Replay &  True &  True\\
 \hline

  \multicolumn{3}{|c|}{\textbf{PPO}}\\\hline
 Number of Layers & 2 & 2\\
 Units Per Layer & 16, 32 & 32, 256  \\
 Gamma & 0.97 &  0.99 \\
 Learning Rate & 0.1 & 0.05\\
 Entropy Coeff. & 0.01 & 0.008\\
 Value Func. Coeff. &  0.5 & 0.5\\
\hline
  \multicolumn{3}{|c|}{\textbf{ACER}}\\\hline
 Number of Layers & 2 & 2\\
 Units Per Layer & 16, 256& 32, 256\\
 Gamma & 0.96 & 0.98 \\
 Learning Rate & 0.01 & 0.003]\\
 Entropy Coeff. & 0.1 & 0.5\\
 buffer Size & 5000 & 5000\\
 
 \hline
  \multicolumn{3}{|c|}{\textbf{GAIL}}\\\hline
 Number of Layers & 2 & 2\\
 Units Per Layer & 16, 256& 32, 256\\
 Gamma & 0.93 & 0.99 \\
 Learning Rate & 0.02 & 0.009 \\
 Entropy Coeff. & 0.1 & 0.43\\
 MAx KL Loss &  0.01 & 0.08\\\hline
 
\end{tabular}
\caption{Final architecture for all of our agents.}
\label{tab:finalArchitecturesAll}
\end{table}

\begin{figure*}
\centering
\includegraphics[width=1\columnwidth]{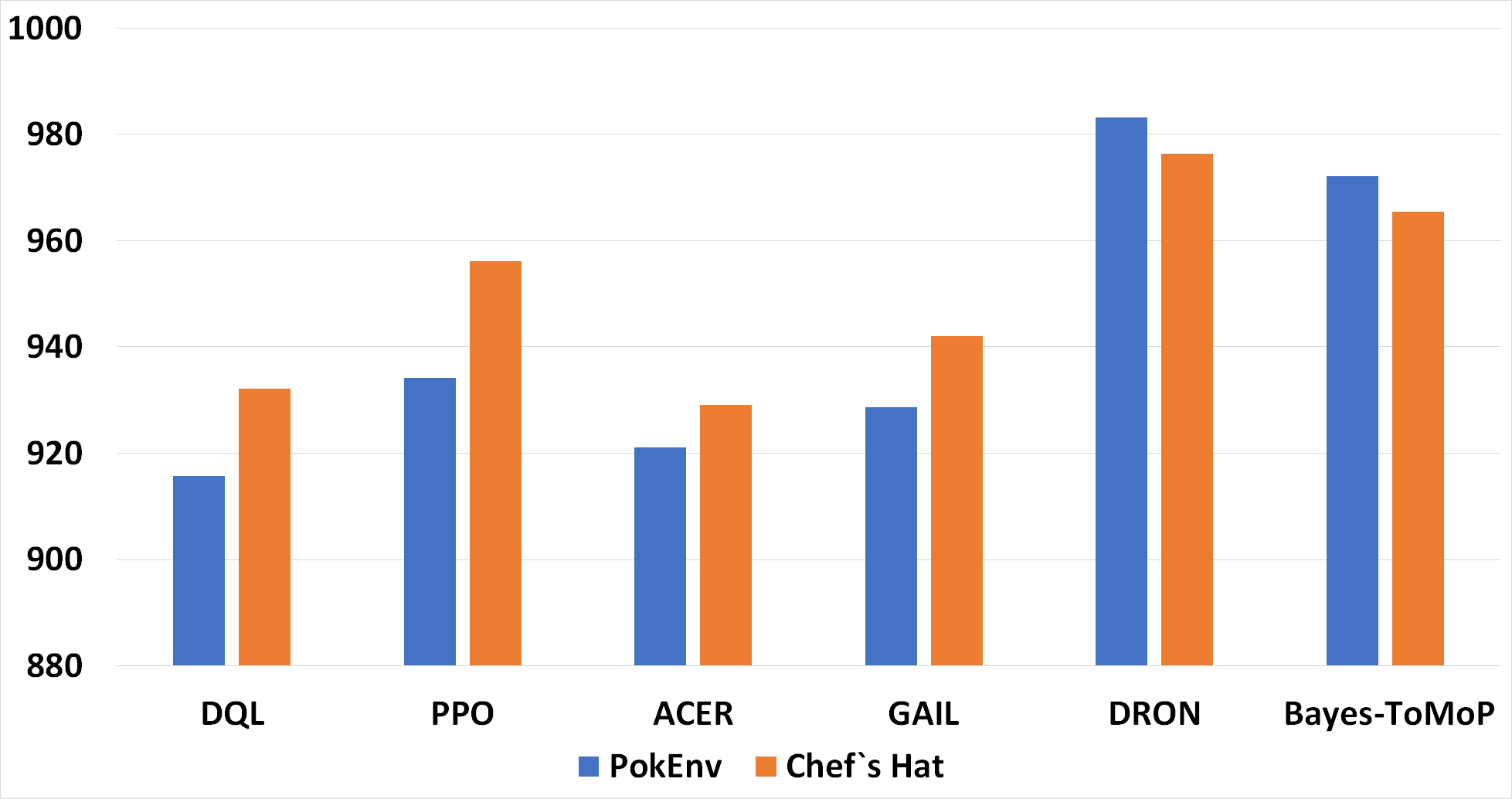}
\caption{Total victories of the best agent of each type after the hyper-parameter selection.}
\label{fig:victoriesModels}
\end{figure*}

The performance of each agent was very similar to each other in both tasks, as exhibited in Figure \ref{fig:victoriesModels}, with the PPO agent reaching the maximum number of 934.2 victories on the PokEnv and 956.2 victories on Chef's Hat.

\textbf{Offline vs Online strategies}

In our tournament, we implement four different versions of each agent: two based on offline training (vs naive and self-playing), and two focusing on online learning (pre-trained and from the scratch).
The \textbf{vs Naive} agents are obtained by training each agent against the simple strategy opponent for 1000 games.
The \textbf{self-playing} strategy has each agent trained with different generations of itself. For each training generation, each agent plays 1000 games. To avoid over-specification of the agents, after each generation we select the best and second-best agents, based on their victory numbers, and add them to an opponent pool. For every generation, we complement the opponent pool with two naive agents, one with a simple strategy and one random, and a newly instantiated agent of that type, without any training. We then make the agent play against randomly selected opponents from the opponent's pool. We run the game for 50 generations and select the best agent of the last generation for our baseline results. The GAIL agent collects the observations of each candidate on the opponent pool to be used as expert observations on the next generation.

The online learning agents learn while playing the game on the experimental tournament. The \textbf{from the scratch} instantiate a new version of the agent, without any prior-knowledge, while the \textbf{pre-trained} one uses the self-play version of the agent as an initial state.

\subsection{CSP implementation}

We employ a PPO agent as both global and local policy, thus, we implement them using the same parameters as mentioned previously.

We optimize the CSP network also using Hyperopt, and the parameters and search space are listed in Table \ref{tab:searchSpaceCPC}. We run 1000 evaluations, and 1000 games for each. We selected the best parameters based on the network's loss, which is optimized by an ADAM optimizer. 

\begin{table}[h!]
\center
\begin{tabular}{ |c|c|} 
\hline
 \textbf{Parameter} & \textbf{Search Space} \\ \hline
 \multicolumn{2}{|c|}{\textbf{Encoder}}\\\hline
 Sequence Size & [3,5,10, 15]\\
 Number of Layers & [1,2,3]\\
 Units Per Layer & [16, 32, 64, 128, 256, 512, 1024] \\
 \hline
  \multicolumn{2}{|c|}{\textbf{Entangled Representation}}\\\hline
 Units & [8, 16, 32, 64, 128] \\
  \hline

\end{tabular}
\caption{Search space used to optimize the CSP network.}
\label{tab:searchSpaceCPC}
\end{table}

The final parameters for the CSP network are described in Table \ref{tab:finalCPC}.
\begin{table}[h!]
\center
\begin{tabular}{ |c|c|c|} 
\hline
 \textbf{Parameter} & \textbf{PokEnv} & \textbf{Chef's Hat} \\ \hline
 \multicolumn{3}{|c|}{\textbf{Encoder}}\\\hline
 Number of Layers & 2 & 2 \\
 Units Per Layer & 16 & 32 \\

  \hline
  \multicolumn{3}{|c|}{\textbf{Entangled Representation}}\\\hline
  Units & 16 & 16  \\
   \hline
\end{tabular}
\caption{Best parameters of the CSP network.}
\label{tab:finalCPC}
\end{table}

\section{Experiment 1) How to play the game}

\begin{figure*}
\centering
\includegraphics[width=1\columnwidth]{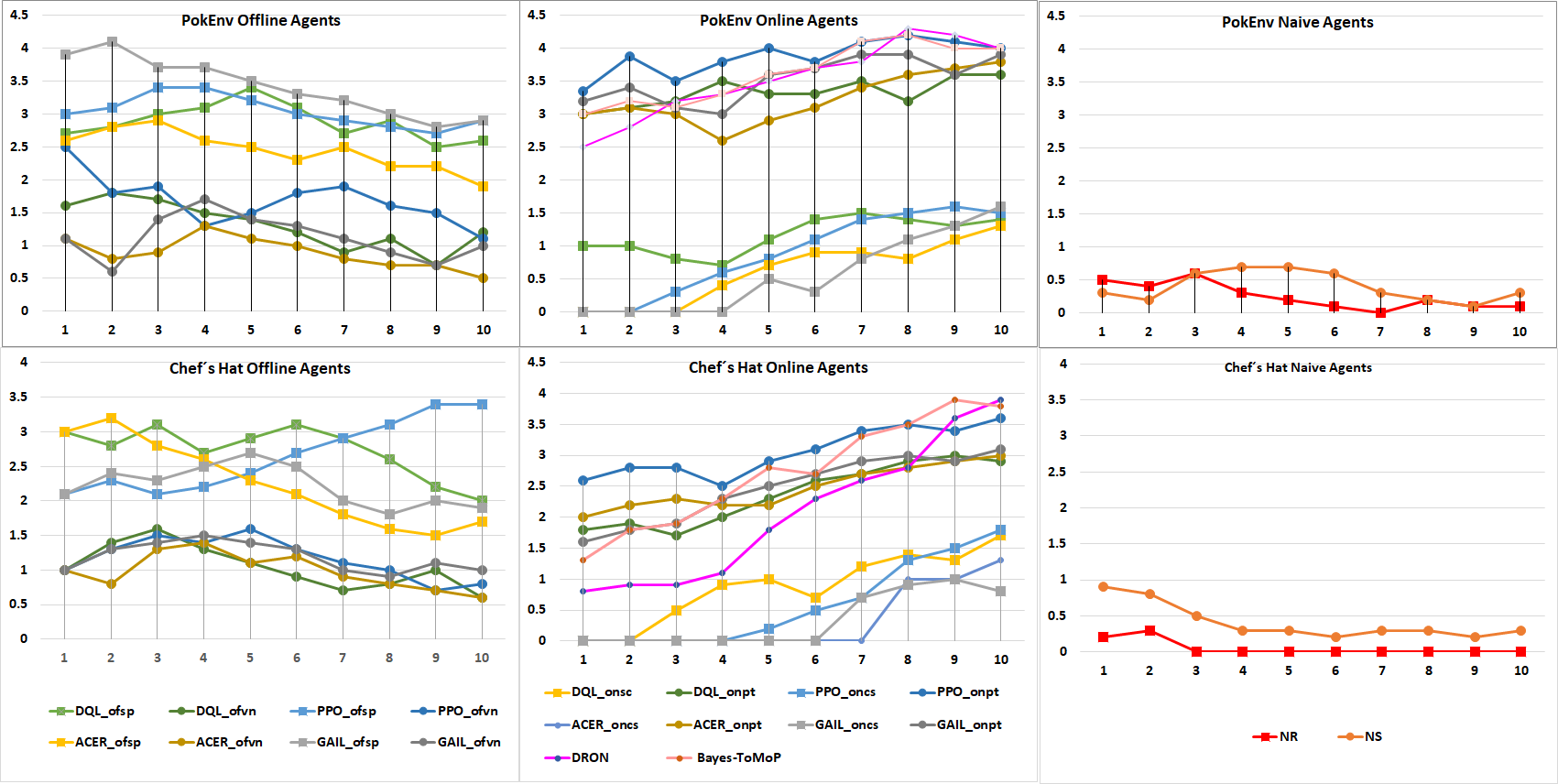}
\caption{Experimental results of all the implemented agents playing 1000 runs of 10 consecutive tournaments for each of the scenarios.}
\label{fig:result1Disc}
\end{figure*}

The entire Experiment 1) results are displayed in Figure \ref{fig:result1Disc}, including the performance of all the offline, online, and naive agents.

\section{Experiment 2) How to predict an opponent's action}

\begin{figure*}
\centering
\includegraphics[width=1\columnwidth]{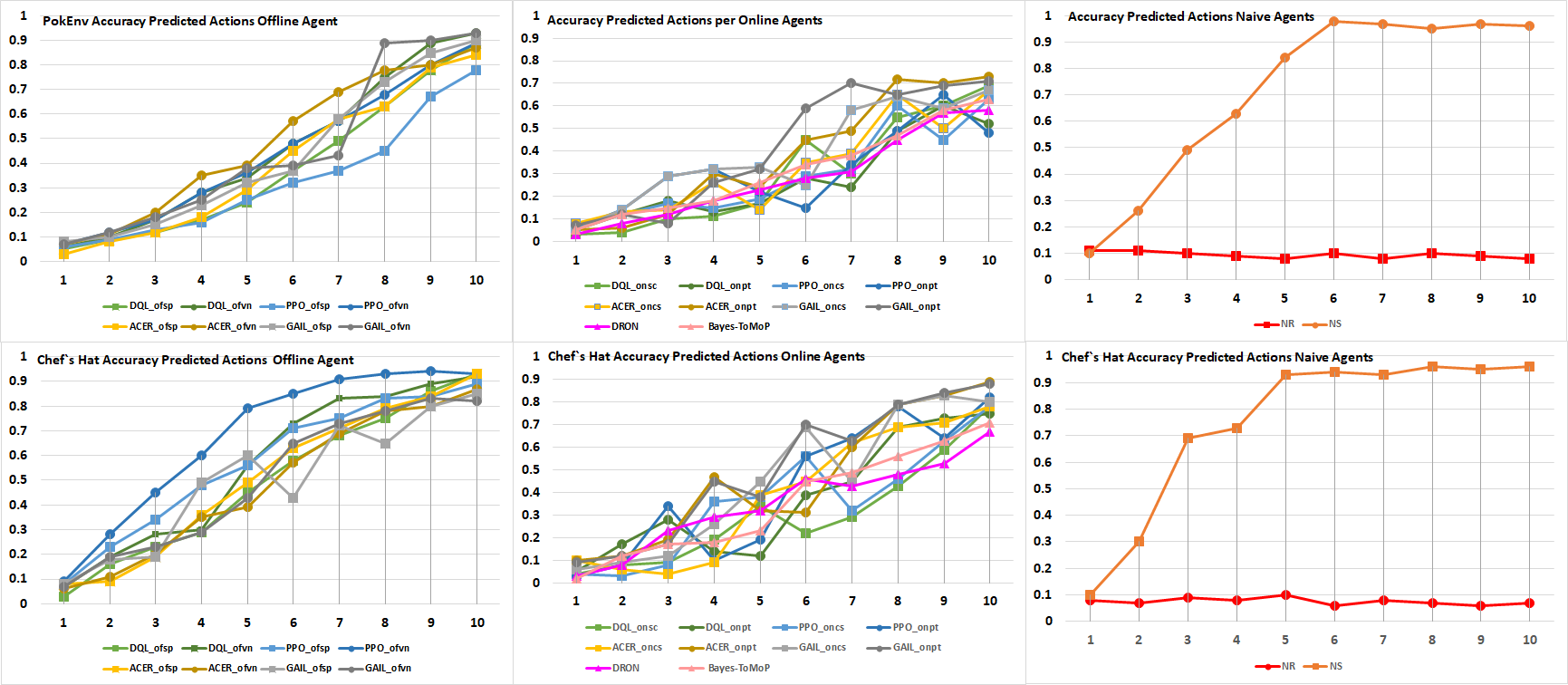}
\caption{Experimental results for our Experiment 2), showing the performance of the CPC network on predicting the opponents' actions.}
\label{fig:result2}
\end{figure*}

The performance of the CPC model to predict the opponents' actions for all the offline, online and naive agents are displayed in Figure \ref{fig:result2}.

\section{Experiment 3) Adapting towards individual opponents}

\begin{figure*}
\centering
\includegraphics[width=1\columnwidth]{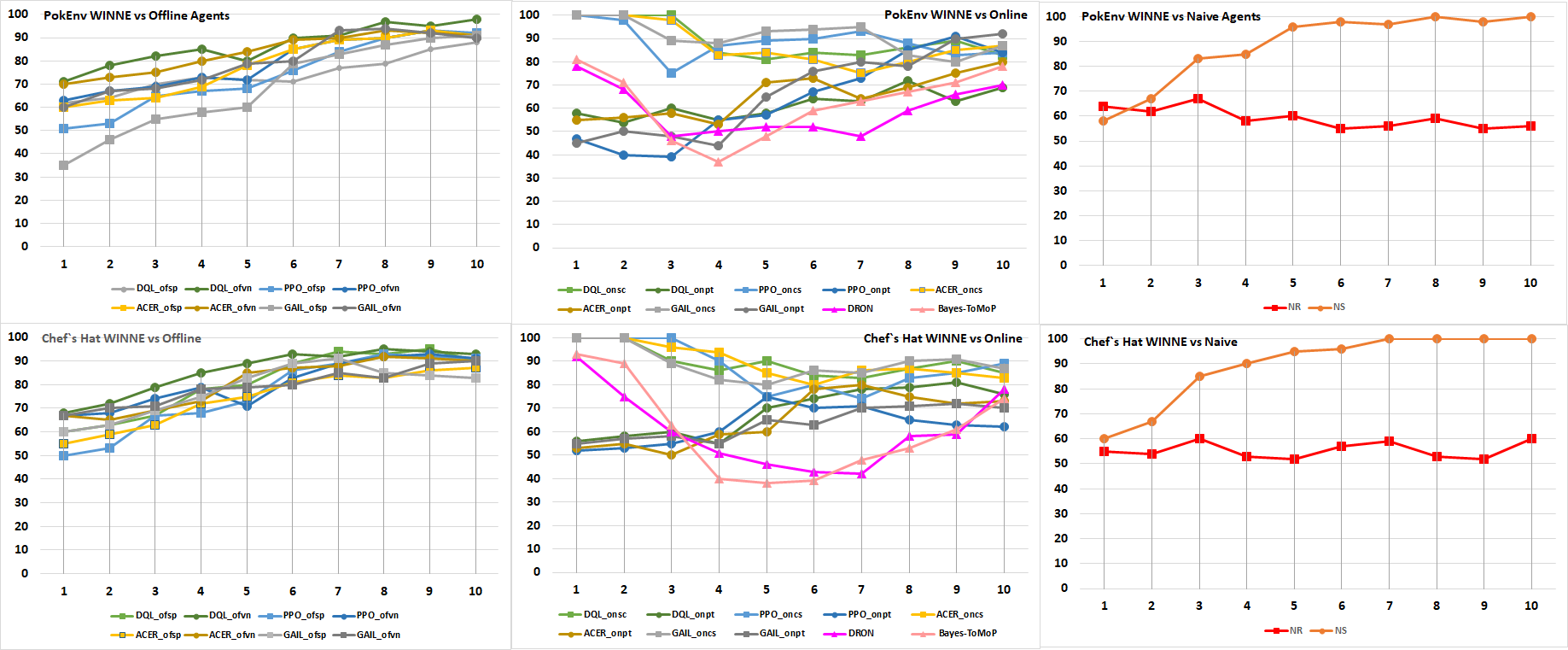}
\caption{Results of Experiment 3), containing the percentage of victories that WINNE has when playing agains all the agents.}
\label{fig:result3}
\end{figure*}

All the results of Experiment 3), containing the performance of WINNE, in terms of percentage of won games, when compared with all the offline, online, and naive agents are displayed in Figure \ref{fig:result3}.

\section{Discussions}

\begin{figure*}
\centering
\includegraphics[width=1\columnwidth]{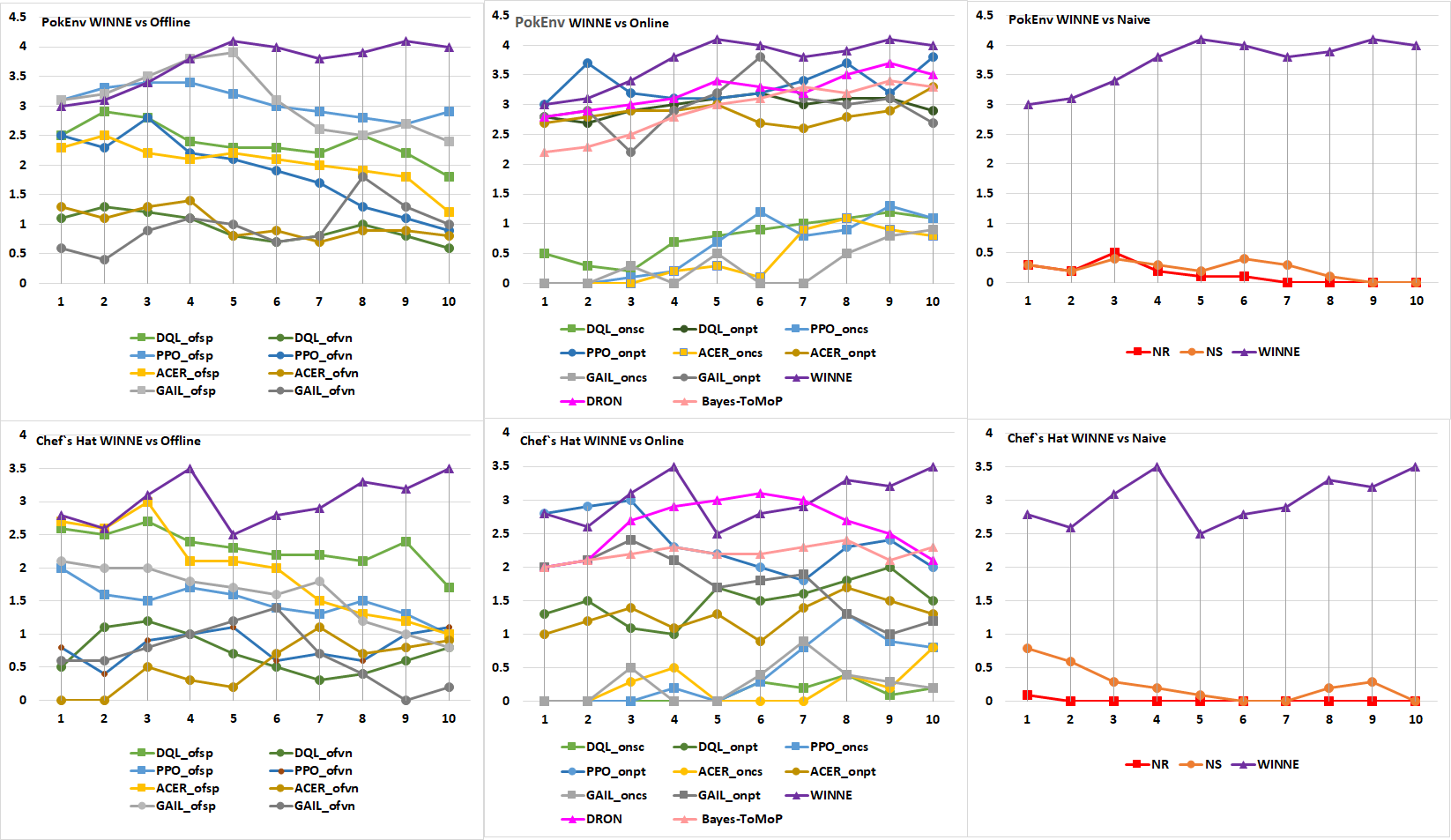}
\caption{Experimental results containing the performance of all of our models when WINNE is present.}
\label{fig:discussion1}
\end{figure*}

All the results of the performance of all agents on the tournament scenario, when WINNE is present, are displayed in Figure \ref{fig:discussion1}.

\end{document}